%% file: www.tex
\documentclass[sigconf]{acmart}

\copyrightyear{2019}
\acmYear{2019}
\setcopyright{iw3c2w3}
\acmConference[WWW '19]{Proceedings of the 2019 World Wide Web Conference}{May 13--17, 2019}{San Francisco, CA, USA}
\acmBooktitle{Proceedings of the 2019 World Wide Web Conference (WWW '19), May 13--17, 2019, San Francisco, CA, USA}
\acmPrice{}
\acmDOI{10.1145/3308558.3313508}
\acmISBN{978-1-4503-6674-8/19/05}

\usepackage{booktabs}
\input{command}

\begin{document}
\title{GraphVite: A High-Performance CPU-GPU Hybrid System for Node Embedding}

\setlength{\belowcaptionskip}{-0.01pt}

\author{Zhaocheng Zhu}
\affiliation{
    \institution{Mila - Qu\'ebec AI Institute}
    \institution{Universit\'e de Montr\'eal}
}
\email{zhaocheng.zhu@umontreal.ca}

\author{Shizhen Xu}
\affiliation{
    \institution{Tsinghua University}
}
\email{xsz12@mails.tsinghua.edu.cn}

\author{Meng Qu}
\affiliation{
    \institution{Mila - Qu\'ebec AI Institute}
    \institution{Universit\'e de Montr\'eal}
}
\email{meng.qu@umontreal.ca}

\author{Jian Tang}
\affiliation{
    \institution{Mila - Qu\'ebec AI Institute}
    \institution{HEC Montr\'eal}
    \institution{CIFAR AI Research Chair}
}
\email{jian.tang@hec.ca}

\renewcommand{\shortauthors}{Zhu et al.}

\begin{abstract}

Learning continuous representations of nodes is attracting growing interest in both academia and industry recently, due to their simplicity and effectiveness in a variety of applications. Most of existing node embedding algorithms and systems are capable of processing networks with hundreds of thousands or a few millions of nodes. However, how to scale them to networks that have tens of millions or even hundreds of millions of nodes remains a challenging problem. In this paper, we propose \Graphy, a high-performance CPU-GPU hybrid system for training node embeddings, by co-optimizing the algorithm and the system. On the CPU end, augmented edge samples are parallelly generated by random walks in an online fashion on the network, and serve as the training data. On the GPU end, a novel parallel negative sampling is proposed to leverage multiple GPUs to train node embeddings simultaneously, without much data transfer and synchronization. Moreover, an efficient collaboration strategy is proposed to further reduce the synchronization cost between CPUs and GPUs. Experiments on multiple real-world networks show that \Graphy is super efficient. It takes only about one minute for a network with 1 million nodes and 5 million edges on a single machine with 4 GPUs, and takes around 20 hours for a network with 66 million nodes and 1.8 billion edges. Compared to the current fastest system, \Graphy is about 50 times faster without any sacrifice on performance.

\end{abstract}

%
%



\maketitle

\input{1_introduction}
\input{3_preliminary}
\input{4_proposed_method}
\input{5_experiment}
\input{6_ablation}
\input{2_related_work}
\input{7_conclusion}
\input{8_acknowledgement}

\bibliographystyle{ACM-Reference-Format}
\bibliography{bibliography}

\end{document}

%% file: command.tex
\usepackage{xspace}
\usepackage{adjustbox}
\usepackage{multirow}
\usepackage{xcolor, colortbl}
\usepackage{booktabs}
\usepackage{dblfloatfix}
\usepackage{graphicx}
\usepackage{subcaption}
\usepackage{amsmath}
\usepackage{amsfonts}
\usepackage{amsthm}
\usepackage{algorithm}
\usepackage{algpseudocode}
\usepackage{enumitem}

\newcommand{\dataset}[1]{\textsc{#1}\xspace}
\newcommand{\mat}[1]{\mathbf{#1}}
\newcommand{\field}[1]{\mathbb{#1}}
\newcommand{\property}[1]{\textit{#1}}
\newcommand{\best}[1]{\textbf{#1}}

\newcommand{\Graphy}{\textit{GraphVite}\xspace}

\newtheorem{definition}{Definition}

%% file: 1_introduction.tex
\section{Introduction}

Networks are ubiquitous in the real world. Examples like social networks \cite{mislove2007measurement}, citation networks \cite{sen2008collective}, protein-protein interaction networks \cite{szklarczyk2016string} and many more cover a wide range of applications. In network analysis, it is critical to have effective representations for nodes, as these representations largely determine the performance of many downstream tasks. Recently, there is a growing interest in unsupervised learning of continuous node representations, which is aimed at preserving the structure of networks in a low-dimensional space. This kind of approaches has been proven successful in various applications, such as node classification \cite{perozzi2014deepwalk}, link prediction \cite{liben2007link}, and network visualization \cite{tang2016visualizing}.

Many works have been proposed on this stream, including DeepWalk \cite{perozzi2014deepwalk}, LINE \cite{tang2015line}, and node2vec \cite{grover2016node2vec}. These methods learn effective node embeddings by predicting the neighbors of each node and can be efficiently optimized by asynchronous stochastic gradient descent (ASGD) \cite{recht2011hogwild}. On a single machine with multi-core CPUs, they are capable of processing networks with one or a few millions of nodes. Given that real-world networks easily go to tens of millions nodes and nearly billions of edges, how to adapt node embedding methods to networks of such large scales remains very challenging. One may think of exploiting computer clusters for training large-scale networks. However, it is a non-trivial task to extend existing methods to distributed settings. Even if distributed algorithms are available, the cost of large CPU clusters is still prohibitive for many users. Therefore, we are wondering whether it is possible to scale node embedding methods to very large networks on a single machine, which should be particularly valuable for common users.

Inspired by the recent success of training deep neural networks with GPUs \cite{ciresan2011flexible, krizhevsky2012imagenet}, we would like to utilize such highly parallel hardware to accelerate the training of node embeddings. However, directly adopting GPUs for node embedding could be inefficient, since the sampling procedure in node embedding requires excessive random memory access on the network structure, which is at the disadvantage of GPUs. Compared to GPUs, CPUs are much more capable of performing random memory access. Therefore, it would be wise to use both CPUs and GPUs for training node embeddings. Along this direction, a straightforward solution is to follow the mini-batch stochastic gradient descent (mini-batch SGD) paradigm utilized in existing deep learning frameworks (e.g. TensorFlow \cite{abadi2016tensorflow} and PyTorch \cite{paszke2017automatic}). Different from deep neural networks, the training of node embeddings involves much more memory access per computation. As a result, mini-batch SGD would suffer from severe memory latency on the bus before it benefits from fast GPU computation. Therefore, other than mini-batch SGD, we need to design a system that leverages distinct advantages of CPUs and GPUs and uses them collaboratively to train node embeddings efficiently.

Overall, the main challenges of building an efficient node embedding system with GPUs are:

\begin{enumerate}[leftmargin=*]
    \item{\textbf{Limited GPU Memory} The parameter matrices of node embeddings are quite large while the memory of a single GPU is very small. Modern GPUs usually have a capacity of 12GB or 16GB. }
    \item{\textbf{Limited Bus Bandwidth} The bandwidth of the bus is much slower than the computation speed of GPUs. There will be severe latency if GPUs exchange data with the main memory frequently. }
    \item{\textbf{Large Synchronization Cost} A lot of data are transferred between CPUs and GPUs. Both the CPU-GPU or inter-GPU synchronizations are very costly.}
\end{enumerate}

In this paper, we propose a high-performance CPU-GPU hybrid system called \Graphy for training node embeddings on large-scale networks. \Graphy takes full advantages of CPUs and GPUs by co-optimizing the node embedding algorithm and the system. Specifically, we observe that existing node embedding methods typically consist of two stages, i.e. network augmentation and embedding training. In the first stage, an augmented network is constructed by random walks on the original network, and positive edges are sampled on the augmented network. In the second stage, node embeddings are trained according to the samples generated in the previous stage. Since the augmentation stage involves excessive random access, we resort to CPUs for this part. The training stage is assigned to GPUs as it is mainly composed of matrix computation.

In \Graphy, the above challenges are addressed by three components, namely parallel online augmentation, parallel negative sampling and collaboration strategy. In parallel online augmentation, CPUs augment the network with random walks and generate edge samples in an online fashion. In parallel negative sampling, the edge samples are organized into a grid sample pool, where each block corresponds to a subset of the network. Then GPUs iteratively fetch orthogonal blocks and their corresponding embeddings in each episode. Because GPUs do not share any embeddings, multiple GPUs can perform gradient updates with negative sampling in its own subset simultaneously. With such a design, the problem of limited GPU memory is solved as each GPU only stores the subset of node embeddings corresponding to the current sample block. The problem of limited bus bandwidth is mitigated since model parameters are transferred only when GPUs change their blocks. No inter-GPU synchronization is needed and CPU-GPU synchronization is only needed at the end of each episode. The collaboration strategy further reduces the synchronization cost between CPUs and GPUs on the sample pool.

We evaluate \Graphy on 4 real-world networks of different scales. On a single machine with 4 Tesla P100 GPUs, our system only takes one minute to train a network with 1 million nodes and 5 million edges. Compared to the current fastest system \cite{tang2015line}, \Graphy is 51 times faster and does not sacrifice any performance. On a network with 66 million nodes and 1.8 billion edges, \Graphy takes only around 20 hours to finish training. We also investigate the speed of \Graphy under different hardware configurations. Even on economic GPUs like GeForce GTX 1080, \Graphy is able to achieve a speedup of 29 times compared to the current fastest system.

\noindent \textbf{Organization} Section \ref{sec:preliminary} reviews existing state-of-the-art node embedding methods and points out the challenges of extending these methods to GPUs. Section \ref{sec:method} introduces our proposed system in details. We present our experiments in Section \ref{sec:experiment}, followed by extensive ablation studies in Section \ref{sec:ablation}. Section \ref{sec:related} summarizes the related work, and we conclude this paper in Section \ref{sec:conclusion}.

%% file: 3_preliminary.tex
\section{Preliminaries}
\label{sec:preliminary}

In this section, some preliminary knowledge is introduced. We first review existing state-of-the-art node embedding methods, followed by a discussion on the main challenges of extending these methods to GPUs.

\subsection{Node Embedding Methods Review}
\label{sec:method_review}

Given a network $G = (V, E)$, the goal of node embedding is to learn a low-dimensional representation for each node. The learned embeddings are expected to capture the structure of the network. Towards this goal, most existing methods train node embeddings to distinguish the edges in $E$ (i.e. positive edges) from some randomly sampled node pairs (i.e. negative edges). In other words, edges are essentially utilized as training data. Since many real-world networks are extremely sparse, most existing embedding methods conduct random walks on the original network to introduce more connectivity. Specifically, they connect nodes within a specified distance on a random walk path as additional positive edges. For example, LINE \cite{tang2015line} uses a breadth-first search strategy on low-degree nodes, while DeepWalk \cite{perozzi2014deepwalk} uses a depth-first search strategy for all nodes. Node2vec \cite{grover2016node2vec} developed a mixture of the above two strategies.

Once the network is augmented, node embeddings are trained on samples from the augmented network. Typically, the embeddings are encoded in two sets, namely $\mat{vertex}$ embedding matrix and $\mat{context}$ embedding matrix. For an edge sample $(u, v)$, the dot product of $\mat{vertex}[u]$ and $\mat{context}[v]$ is computed to predict whether the sample is a positive edge. This encourages neighbor nodes to have close embeddings, whereas distant nodes will have very different embeddings.

Overall, the computation procedures of these node embedding methods can be divided into two stages: \textbf{network augmentation} and \textbf{embedding training}. Algorithm \ref{alg:cpu} summarizes the general framework of existing node embedding methods. Note that the first stage can be easily parallelized, and the second stage can be parallelized via asynchronous SGD. In most existing node embedding systems, these two stages are executed in a sequential order, with each stage parallelized by a bunch of CPU threads.

\begin{algorithm}[!h]
    \caption{General framework of node embedding}
    \begin{algorithmic}[1]
        \State{$E' \gets E$}
        \For{$v \in V$} \Comment {parallelizable}
            \For{$u \in \Call{Walk}{v}$}
                \State{$E' \gets E' \cup \{(v, u, \Call{Weight}{v, u})\}$}
            \EndFor
        \EndFor
        
        \State{}
        
        \For{each iteration} \Comment {parallelizable}
            \State{$v, u \gets \Call{EdgeSampling}{E'}$}
            \State{$\Call{Train}{\mat{vertex}[v], \mat{context}[u], label=1}$}
            \For{$u' \in \Call{NegativeSampling}{V}$}
                \State{$\Call{Train}{\mat{vertex}[v], \mat{context}[u'], label=0}$}
            \EndFor
        \EndFor
    \end{algorithmic}
    \label{alg:cpu}
\end{algorithm}

\subsection{Challenges for Hybrid Node Embedding System}
\label{sec:challenges}

Inspired by the recent success of training neural networks with GPUs, we are interested in building a node embedding system by leveraging the power of GPUs, which can benefit the embedding training stage. Since the first stage of network augmentation involves extensive memory random access, CPUs are more suitable for this stage. As a result, we desire to develop a hybrid CPU-GPU system for training node embeddings. A common approach for a hybrid machine learning system is the mini-batch SGD paradigm, which is widely adopted in existing deep learning frameworks such as TensorFlow \cite{abadi2016tensorflow} and PyTorch \cite{paszke2017automatic}. In mini-batch SGD, model parameters are stored on GPUs and training data is iteratively passed to GPUs in batches.

However, mini-batch SGD cannot be applied directly to node embedding on large networks. Take a scale-free network with 50 million nodes and 1 billion edges as an example. (1) The size of the augmented network goes to 373 GB large, which may overwhelm the memory of most servers. We need to figure out a way to generate the augmented network and edge samples on the fly. (2) The embedding matrices are much larger than the parameter matrices in deep neural networks. Either $\mat{vertex}$ or $\mat{context}$ matrix consumes 23.8 GB memory, which is beyond the memory limit of any single GPU. As a result, both the $\mat{vertex}$ and $\mat{context}$ embedding matrices have to be stored in the main memory, and transferred to the GPUs in small parts during training. See Table \ref{tab:memory_cost} for a detailed analysis of the memory cost. 

\begin{table}[h]
    \centering
    \begin{adjustbox}{max width=0.48\textwidth}
        \begin{tabular}{lccc}
            \toprule
                                & Size              & Example               & Minimum storage   \\
            \midrule
            nodes               & $|V|$             & $5*10^7$              & 191 MB            \\
            edges               & $|E|$             & $1*10^9$              & 7.45 GB           \\
            augmented edges     & $|E'|$            & $5*10^{10}$           & 373 GB            \\
            $\mat{vertex}$      & $|V| \times d$    & $5*10^7 \times 128$   & 23.8 GB           \\
            $\mat{context}$     & $|V| \times d$    & $5*10^7 \times 128$   & 23.8 GB           \\
            \bottomrule
        \end{tabular}
    \end{adjustbox}
    \caption{Memory cost of node embedding on a scale-free network with 50 million nodes and 1 billion edges.}
    \label{tab:memory_cost}
\end{table}

For the second problem, while transferring parameters from CPUs and GPUs sounds feasible, it will become a bottleneck if we consider the bus bandwidth between CPUs and GPUs. Consider training $d$-dimensional embeddings with $n$ edge samples. There is $O(nd)$ computation workload and also $O(nd)$ memory access if the samples are not overlapped with each other. Since the computation speed of GPUs is way faster than the speed of bus transfer, the entire system will be bounded by the speed of parameter transfer from CPUs to GPUs severely. Indeed, such a system is even worse than its CPU parallel counterpart, which is verified in our experiments (see Table \ref{tab:time_youtube}).

Another challenge in a hybrid system is the large synchronization cost. Since the system is distributed on multiple CPUs and GPUs, there is necessary data (e.g. parameters and edge samples) shared across sub tasks. A trivial but safe solution is to synchronize the shared data frequently, which will result in huge synchronization cost. To achieve high speed performance, the system should reduce shared data as much as possible, and use a collaboration strategy to minimize synchronization cost between devices.

%% file: 4_proposed_method.tex
\section{\Graphy: a Hybrid CPU-GPU Node Embedding System}
\label{sec:method}

\begin{figure}[!h]
    \centering
    \includegraphics[width=0.48\textwidth]{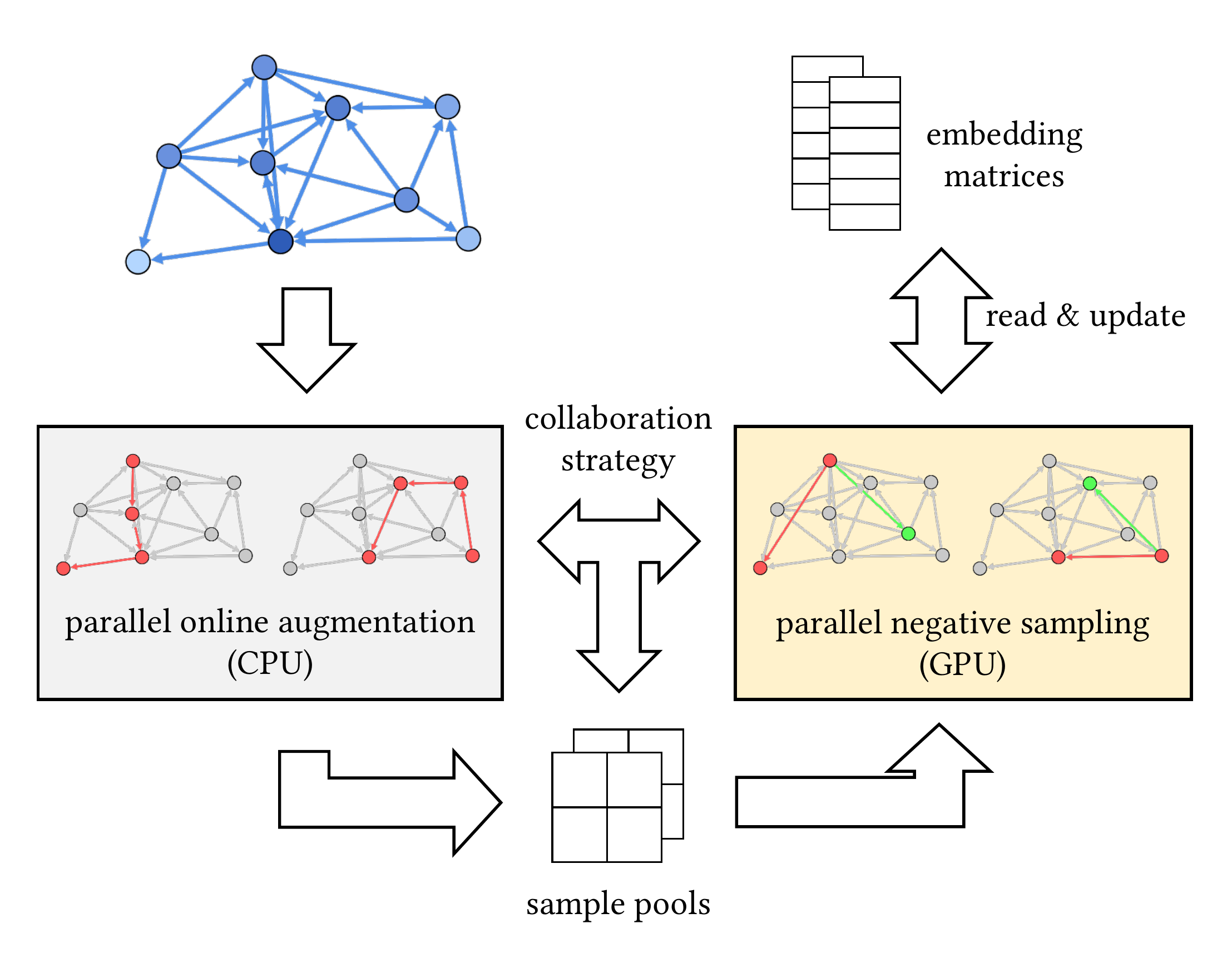}
    \caption{Overview of our hybrid system. The gray and yellow boxes correspond to the stages of network augmentation and embedding training respectively. The former is performed by parallel online augmentation on CPUs, while the latter is performed by parallel negative sampling on GPUs. The two stages are executed asynchronously with our collaboration strategy.}
    \label{fig:system}
\end{figure}

In this section, we introduce a high-performance hybrid CPU-GPU system called \Graphy for training node embeddings. Our system leverages distinct advantages of CPUs and GPUs and addresses the above three challenges. Specifically, we propose a parallel online augmentation for efficient network augmentation on CPUs. We introduce a parallel negative sampling to cooperate multiple GPUs for embedding training. A collaboration strategy is also proposed to reduce the synchronization cost between CPUs and GPUs.

\subsection{Parallel Online Augmentation}
\label{sec:parallel_augmentation}

\begin{figure*}
    \includegraphics[width=0.95\textwidth]{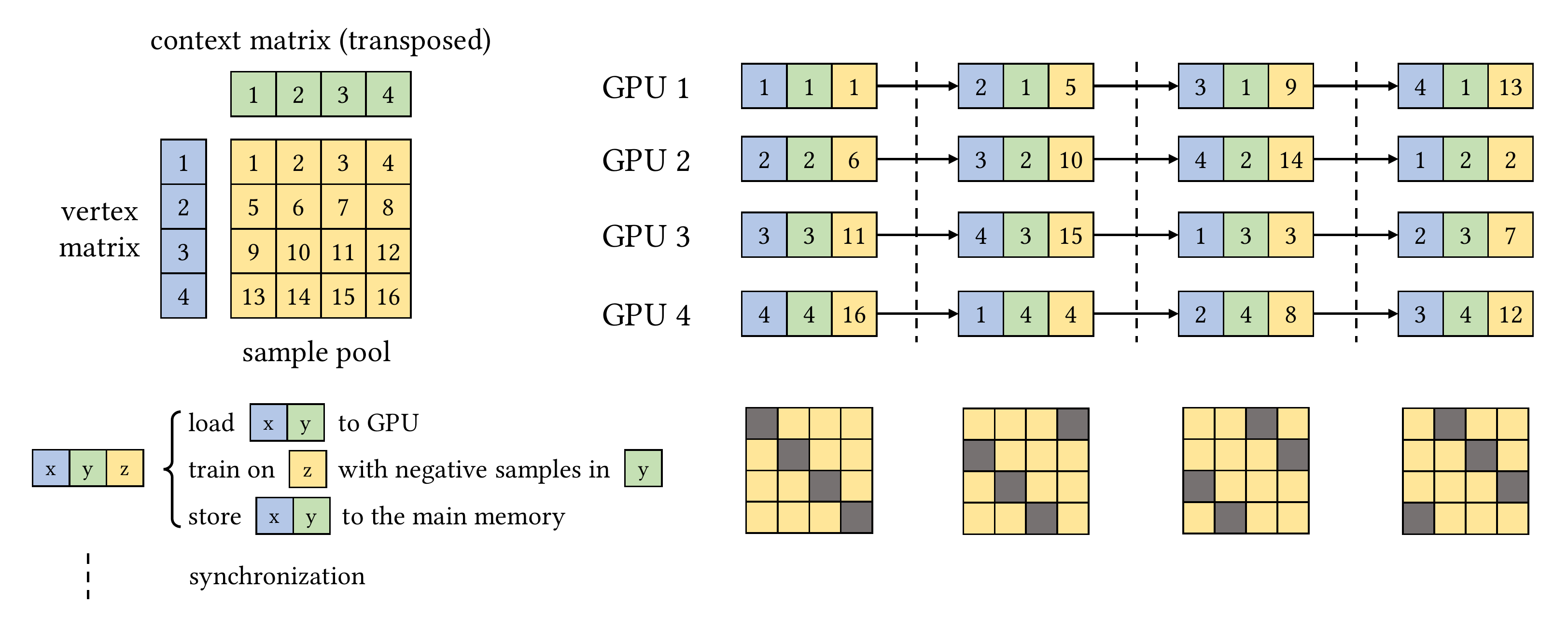}
    \caption{Illustration of parallel negative sampling on 4 GPUs. During each episode, GPUs take orthogonal blocks from the sample pool. Each GPU trains embeddings with negative samples drawn from its own context nodes. Synchronization is only needed between episodes.}
    \label{fig:parallel_negative_sampling}
\end{figure*}

As discussed in Section \ref{sec:method_review}, the first stage of node embedding methods is to augment the original network with random walks. Since the augmented network is usually one or two magnitude larger than the original one, it is impossible to load it into the main memory if the original network is already very large. Therefore, we introduce a parallel online augmentation, which generates augmented edge samples on the fly without explicit network augmentation. Our method can be viewed as an online extension of the augmentation and edge sampling method used in LINE\cite{tang2015line}. First, we draw a departure node with the probability proportional to the degree of each node. Then we perform a random walk from the departure node, and pick node pairs within a specific augmentation distance $s$ as edge samples. Note that edge samples generated in the same random walk are correlated and may degrade the performance of optimization. Inspired by the experience replay technique widely used in reinforcement learning \cite{lin1993reinforcement, mnih2013playing}, we collect edge samples into a sample pool, and shuffle the sample pool before transferring it to GPUs for embedding training. The proposed edge sampling method can be parallelized when each thread is allocated with an independent sample pool in advance. Algorithm \ref{alg:parallel_online_augmentation} gives the process of parallel online augmentation in details.

\begin{algorithm}
    \caption{Parallel Online Augmentation}
    \begin{algorithmic}[1]
        \Function{ParallelOnlineAugmentation}{$num\_CPU$}
            \For{$i \gets 0$ to $num\_CPU - 1$} \Comment{paralleled}
                \State{$pool[i] \gets \varnothing$}
                \While{$pool$ is not full}
                    \State{$x \gets \Call{DepartureSampling}{G}$}
                    \For{$u, v \in \Call{RandomWalkSampling}{x}$}
                        \If{$Distance(u, v) <= s$}
                            \State{$pool.append((u, v))$}
                        \EndIf
                    \EndFor
                \EndWhile
                \State{$pool[i] \gets \Call{Shuffle}{pool[i]}$}
            \EndFor
            \State{\Return${\Call{Concatenate}{pool[\cdot]}}$}
        \EndFunction
    \end{algorithmic}
    \label{alg:parallel_online_augmentation}
\end{algorithm}

\noindent \textbf{Pseudo Shuffle}
While shuffling the sample pool is important to optimization, it slows down the network augmentation stage (see Table \ref{tab:shuffle}). The reason is that a general shuffle consists of lots of random memory access and cannot be accelerated by the CPU cache. The loss in speed will be even worse if the server has more than one CPU socket. To mitigate this issue, we propose a pseudo shuffle technique that shuffles correlated samples in a much more cache-friendly way and improves the speed of the system significantly. Note that most correlation comes from edge samples that share the source node or the target node in the same random walk. As such correlation occurs in a group of $s$ samples for an augmentation distance $s$, we divide the sample pool into $s$ continuous blocks, and scatter correlated samples into different blocks. For each block, we always append samples sequentially at the end, which can benefit a lot from CPU cache. The $s$ blocks are concatenated to form the final sample pool.

\subsection{Parallel Negative Sampling}
\label{sec:parallel_training}

In the embedding training stage, we divide the training task into small fragments and distribute them to multiple GPUs. The sub tasks are necessarily designed with little shared data to minimize the synchronization cost among GPUs. To see how model parameters can be distributed to multiple GPUs without overlap, we first introduce a definition of \property{$\epsilon$-gradient exchangeable}.

\begin{definition}
    \textbf{$\epsilon$-gradient exchangeable}. A loss function $L(X;\theta)$ is \property{$\epsilon$-gradient exchangeable} on two sets of training data $X_1$, $X_2$ if for small $\epsilon \geq 0$, $\forall \theta_0 \in \Theta$ and $\forall \alpha \in \field{R^+}$, exchanging the order of two gradient descent steps results in a vector difference with norm no more than $\epsilon$.
    \begin{equation}
        \begin{cases}
            & \theta_1 \gets \theta_0 - \alpha \nabla L(X_1;\theta_0) \\
            & \theta_2 \gets \theta_1 - \alpha \nabla L(X_2;\theta_1)
        \end{cases}
        \label{eq:order1}
    \end{equation}
    \begin{equation}
        \begin{cases}
            & \theta'_1 \gets \theta_0 - \alpha \nabla L(X_2;\theta_0) \\
            & \theta'_2 \gets \theta'_1 - \alpha \nabla L(X_1;\theta'_1)
        \end{cases}
        \label{eq:order2}
    \end{equation}
    i.e. $\lVert \theta_2 - \theta'_2 \rVert \leq \epsilon$ is true for the above equations.
\end{definition}

Particularly, we abbreviate \property{0-gradient exchangeable} to \property{gradient exchangeable}. Due to the sparse nature of node embedding training, there are many sets that form \property{gradient exchangeable} pairs in the network. For example, for two edge sample sets $X_1, X_2 \subseteq E$, if they do not share any source nodes or target nodes, $X_1$ and $X_2$ are \property{gradient exchangeable}. Even if $X_1$ and $X_2$ share some nodes, they can still be \property{$\epsilon$-gradient exchangeable} if the learning rate $\alpha$ and the number of iterations are bounded.

Based on the gradient exchangeability observed in node embedding, we propose a parallel negative sampling algorithm for the embedding training stage. For $n$ GPUs, we partition rows of $\mat{vertex}$ and $\mat{context}$ into $n$ partitions respectively (see the top-left corner of Figure \ref{fig:parallel_negative_sampling}). This results in an $n \times n$ partition grid for the sample pool, where each edge belongs to one of the blocks. In this way, any pair of blocks that does not share row or column is \property{gradient exchangeable}. Blocks in the same row or column are \property{$\epsilon$-gradient exchangeable}, as long as we restrict the number of iterations on each block.

We define \emph{episode} as the block-level step used in parallel negative sampling. During each episode, we send $n$ orthogonal blocks and their corresponding $\mat{vertex}$ and $\mat{context}$ partitions to $n$ GPUs respectively. Each GPU then updates its own embedding partitions with ASGD. Because these blocks are mutually \property{gradient exchangeable} and do not share any row in the parameter matrices, multiple GPUs can perform ASGD concurrently without any synchronization. At the end of each episode, we gather the updated parameters from all GPUs and assign another $n$ orthogonal blocks. Here \property{$\epsilon$-gradient exchangeable} is controlled by the number of total samples in $n$ orthogonal blocks, which we define as \emph{episode size}. The smaller episode size, the better \property{$\epsilon$-gradient exchangeable} we will have for embedding training. However, smaller episode size will also induce more frequent synchronization. Hence the episode size is tuned so that there is a good trade off between the speed and \property{$\epsilon$-gradient exchangeable} (see Section \ref{sec:exchangeability}). Figure \ref{fig:parallel_negative_sampling} gives an example of parallel negative sampling with 4 partitions.

Typically, node embedding methods sample negative edges from all possible nodes. However, it could be very time-consuming if GPUs have to communicate with each other to get the embeddings of their negative samples. To avoid this cost, we restrict that negative samples can only be drawn from the $\mat{context}$ rows on the current GPU. Though this seems a little problematic, we find it works well in practice. An intuitive explanation is that with parallel online augmentation, every node is likely to have positive samples with nodes from all context partitions. As a result, every node can potentially form negative samples with all possible nodes.

Note that although we demonstrate with the number of partitions equal to $n$, the parallel negative sampling can be easily generalized to cases with any number of partitions greater than $n$, simply by processing the orthogonal blocks in subgroups of $n$ during each episode. Algorithm \ref{alg:parallel_negative_sampling} illustrates the hybrid system for multiple GPUs.

\begin{algorithm}
    \caption{Parallel Negative Sampling}
    \begin{algorithmic}[1]        
        \Function{ParallelNegativeSampling}{$num\_GPU$}        
            \State{$vertex\_partitions \gets \Call{Partition}{\mat{vertex}}$}
            \State{$context\_partitions \gets \Call{Partition}{\mat{context}}$}
            \While{not converge}
                \State{$pool \gets \Call{ParallelOnlineAugmentation}{num\_CPU}$}
                \State{$block[\cdot][\cdot] \gets \Call{Redistribute}{pool}$}
                \For{$\mathit{offset} \gets 0$ to $num\_GPU - 1$}
                    \For{$i \gets 0$ to $num\_GPU - 1$} \Comment{paralleled}
                        \State{$vid \gets i$}
                        \State{$cid \gets (i + \mathit{offset}) \mod num\_GPU$}
                        \State{send $vertex\_partitions[vid]$ to GPU $i$}
                        \State{send $context\_partitions[cid]$ to GPU $i$}
                        \State{train $block[vid][cid]$ on GPU $i$}
                        \State{receive $vertex\_partitions[vid]$ from GPU $i$}
                        \State{receive $context\_partitions[cid]$ from GPU $i$}
                    \EndFor
                \EndFor
            \EndWhile
        \EndFunction
    \end{algorithmic}
    \label{alg:parallel_negative_sampling}
\end{algorithm}

\subsection{Collaboration Strategy}
\label{ref:collaboration_strategy}

Our parallel negative sampling enables different GPUs to train node embeddings concurrently, with only synchronization required between episodes. However, it should be noticed that the sample pool is also shared between CPUs and GPUs. If they synchronize on the sample pool, then only workers of the same stage can access the pool at the same time, which means hardware is idle for half of the time. To eliminate this problem, we propose a collaboration strategy to reduce the synchronization cost. We allocate two sample pools in the main memory, and let CPUs and GPUs always work on different pools. CPUs first fill up a sample pool and pass it to GPUs. After that, parallel online augmentation and parallel negative sampling are performed concurrently on CPUs and GPUs respectively. The two pools are swapped when CPUs fill up a new pool. Figure \ref{fig:system} illustrates this procedure. With the collaboration strategy, the synchronization cost between CPUs and GPUs is reduced and the speed of our hybrid system is almost doubled.

\begin{table*}
    \centering
    \begin{tabular}{ccccc}
        \toprule
        Dataset         & \dataset{Youtube}             & \dataset{Friendster-small}    & \dataset{Hyperlink-PLD}   & \dataset{Friendster}          \\
        \midrule
        $|V|$           & 1,138,499                     & 7,944,949                     & 39,497,204                & 65,608,376                    \\
        $|E|$           & 4,945,382                     & 447,219,610                   & 623,056,313               & 1,806,067,142                 \\
        Evaluation Task & 47-class node classification  & 100-class node classification & link prediction           & 100-class node classification \\
        \bottomrule
    \end{tabular}
    \caption{Statistics of the datasets used in experiments}
    \label{tab:datasets}
\end{table*}

\subsection{Discussion}
Here we further discuss some practical details of our hybrid system.

\smallskip \noindent \textbf{Batched Transfer} In parallel negative sampling, the sample pool is assigned to GPUs by block, which is sometimes very large for the memory of a GPU. Instead of copying the whole sample block to a GPU, we transfer the sample block by a small granularity. In this way, the memory cost of edge samples on GPUs becomes negligible.

\smallskip \noindent \textbf{Bus Usage Optimization} When the number of partitions equals the number of GPUs, we can further optimize the bus usage by fixing the context partition for each GPU. In this way, we save the transfer of $context$ matrix and further reduce the synchronization cost between CPUs and GPUs.

\smallskip \noindent \textbf{Single GPU Case} Although parallel negative sampling is proposed for multiple GPUs, our hybrid system is compatible with a single GPU. Typically a GPU can hold at most 12 million node embeddings. So a single GPU is sufficient for training node embeddings on networks that contain no more than 12 million nodes.

%% file: 5_experiment.tex
\section{Experiments}
\label{sec:experiment}

In this section, we verify the effectiveness and efficiency of \Graphy. We first evaluate our system on \dataset{Youtube}, which is a large network widely used in the literature of node embeddings. Then we evaluate \Graphy on three larger datasets.

\subsection{Datasets}

We use the following datasets in our experiments. Statistics of these networks are summarized in Table \ref{tab:datasets}.

\begin{itemize}[leftmargin=10pt]
    \begin{item}
        \dataset{Youtube} \cite{mislove2007measurement} is a large-scale social network in the Youtube website. It contains 1 million nodes and 5 million edges. For some of the nodes, they have labels that represent the type of videos users enjoy.
    \end{item}
    \begin{item}
        \dataset{Friendster-small} \cite{yang2015defining} is a sub-graph induced by all the labeled nodes in \dataset{Friendster}. It has 8 million nodes and 447 million edges. The node labels in this network are the same as those in \dataset{Friendster}.
    \end{item}
    \begin{item}
        \dataset{Hyperlink-PLD} \cite{meusel2015graph} is a hyperlink network extracted from the Web corpus \footnote{\url{http://commoncrawl.org/}}. We use the pay-level-domain aggregated version of the network. It has 43 million nodes and 623 million edges. This dataset does not contain any label.
    \end{item}
    \begin{item}
        \dataset{Friendster} \cite{yang2015defining} is a very large social network in an online gaming site. It has 65 million nodes and 1.8 billion edges. Some nodes have labels that represent the group users join.
    \end{item}
\end{itemize}

\begin{table*}
    \centering
    \begin{tabular}{lcccc}
         \toprule
         Method                                 & CPU threads   & GPU       & Training time                     & Preprocessing time                \\
         \midrule
         LINE \cite{tang2015line}               & 20            & -         & 1.24 hrs                          & 17.4 mins                         \\
         DeepWalk \cite{perozzi2014deepwalk}    & 20            & -         & 1.56 hrs                          & 14.2 mins                         \\
         node2vec \cite{grover2016node2vec}     & 20            & -         & 47.7 mins                         & 25.9 hrs                          \\
         \midrule
         LINE in OpenNE \cite{thunlp2017openne} & 1             & 1         & > 1 day                           & \colorbox{lightgray}{2.14 mins}   \\
         \midrule
         GraphVite                              & 6             & 1         & \best{3.98 mins}($18.7\times$)    & \best{7.37 s}                     \\
         GraphVite                              & 24            & 4         & \best{1.46 mins}($50.9\times$)    & \best{16.0 s}                     \\
         \bottomrule
    \end{tabular}
    \caption{Results of time of different systems on \dataset{Youtube}. The preprocessing time refers to all the overhead before training, including network input and offline network augmentation. Note the preprocessing time of OpenNE is not comparable since it does not have the network augmentation stage. The speedup ratio of \Graphy is computed with regard to LINE, which is the current fastest system.}
    \label{tab:time_youtube}
\end{table*}

\begin{table*}
    \centering
    \begin{tabular}{llcccccccccc}
        \toprule
                                        & \% Labeled Nodes                      & 1\%           & 2\%           & 3\%           & 4\%           & 5\%
                                                                                & 6\%           & 7\%           & 8\%           & 9\%           & 10\%  \\
        \midrule
        \multirow{4}{*}{Micro-F1(\%)}   & LINE\cite{tang2015line}               & 32.98         & 36.70         & 38.93         & 40.26         & 41.08
                                                                                & 41.79         & 42.28         & 42.70         & 43.04         & 43.34 \\
                                        & LINE\cite{tang2015line}+augmentation  & 36.78         & 40.37         & 42.10         & 43.25         & 43.90
                                                                                & 44.44         & 44.83         & 45.18         & 45.50         & 45.67 \\
                                        & DeepWalk\cite{tang2015line}           & \best{39.68}  & 41.78         & 42.78         & 43.55         & 43.96
                                                                                & 44.31         & 44.61         & 44.89         & 45.06         & 45.23 \\
       \cmidrule{2-12}
                                        & GraphVite                             & 39.19         & \best{41.89}  & \best{43.06}  & \best{43.96}  & \best{44.53}
                                                                                & \best{44.93}  & \best{45.26}  & \best{45.54}  & \best{45.70}  & \best{45.86}  \\ 
        \midrule
        \multirow{4}{*}{Macro-F1(\%)}   & LINE\cite{tang2015line}               & 17.06         & 21.73         & 25.28         & 27.36         & 28.50
                                                                                & 29.59         & 30.43         & 31.14         & 31.81         & 32.32 \\
                                        & LINE\cite{tang2015line}+augmentation  & 22.18         & 27.25         & 29.87         & 31.88         & 32.86
                                                                                & 33.73         & 34.50         & 35.15         & 35.76         & 36.19 \\
                                        & DeepWalk\cite{tang2015line}           & \best{28.39}  & \best{30.96}  & \best{32.28}  & \best{33.43}  & \best{33.92}
                                                                                & 34.32         & 34.83         & 35.27         & 35.54         & 35.86         \\
        \cmidrule{2-12}
                                        & GraphVite                             & 25.61         & 29.46         & 31.32         & 32.70         & 33.81
                                                                                & \best{34.59}  & \best{35.27}  & \best{35.82}  & \best{36.14}  & \best{36.49} \\ 
        \bottomrule
    \end{tabular}
    \caption{Results of node classification on \dataset{Youtube}}
    \label{tab:performance_youtube}
\end{table*}

\subsection{Compared Systems}

We compare \Graphy with the following node embedding systems.

\begin{itemize}[leftmargin=10pt]
    \begin{item}
        LINE \cite{tang2015line} \footnote{\url{https://github.com/tangjianpku/LINE} \label{fn:line_url}} is a CPU parallel system based on C++. We parallel its network augmentation stage for fair comparison with other methods.
    \end{item}
    \begin{item}
        DeepWalk \cite{perozzi2014deepwalk} \footnote{\url{https://github.com/phanein/deepwalk} \label{fn:deepwalk_url}} is a CPU parallel system based on Python and gensim \cite{rehurek2010software}.
    \end{item}
    \begin{item}
        node2vec \cite{grover2016node2vec} \footnote{\url{https://github.com/aditya-grover/node2vec} \label{fn:node2vec_url}} is another CPU parallel system based on Python and gensim \cite{rehurek2010software}.
    \end{item}
    \begin{item}
        LINE in OpenNE \cite{thunlp2017openne} \footnote{\url{https://github.com/thunlp/OpenNE}} is a GPU system based on the Python and TensorFlow \cite{abadi2016tensorflow}.
    \end{item}
\end{itemize}

Although DeepWalk and node2vec are implemented in Python, their computation is fully paralleled by Cython code without GIL, which is comparable to C++ implementations.

\subsection{Implementation Details}

Our implementation generally follows the open source codes of LINE \textsuperscript{\ref{fn:line_url}} and DeepWalk \textsuperscript{\ref{fn:deepwalk_url}}. We adopt the asynchronous SGD \cite{recht2011hogwild} in GPU training, and leverage the on-chip shared memory of GPU for fast forward and backward propagation. We also utilize the alias table trick \cite{tang2015line, grover2016node2vec} to boost parallel online augmentation and parallel negative sampling.

\begin{figure}[!h]
    \centering
    \includegraphics[width=0.35\textwidth]{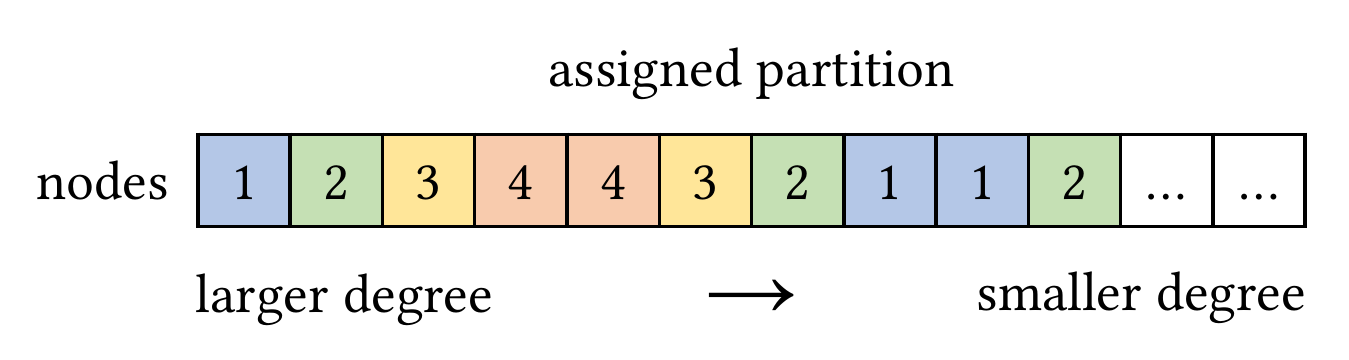}
    \caption{Degree-guided node and context partition strategy in the case of 4 partitions.}
    \label{fig:zig-zag_partition}
\end{figure}

Our hyperparameters are set according to the settings in LINE \cite{tang2015line} and DeepWalk \cite{perozzi2014deepwalk}. We treat networks as undirected graphs. During the network augmentation stage, we sample random walks with a length of 40 edges. We use a degree-guided strategy to partition both $\mat{vertex}$ and $\mat{context}$ matrices. More specifically, we first sort nodes by their degrees and then assign them into different partitions in a zig-zag fashion, as illustrated in Figure \ref{fig:zig-zag_partition}. We tune the episode size to maximize the speed of our hybrid system. During the embedding training stage, negative samples are sampled with a probability proportional to the $3/4$ power of the node degrees. For each positive sample, we draw 1 negative sample and scale the gradient of the negative sample by 5 to match the gradient scale in LINE. We follow the initial learning rate of 0.025 and the linear learning rate decay mechanism in LINE and DeepWalk. We only adopt the \textit{O3} optimization in \textit{g++} and \textit{nvcc}. We do not use any non-standard optimizations or low precision training \cite{zhou2016dorefa, micikevicius2017mixed}, though they may further improve the speed of our system.

We define a training epoch as training $|E|$ positive edge samples. For \dataset{Youtube}, the length of random walk is set to 5, and the total number of training epochs is set to 4,000. For the other 3 datasets, the length of random walks is set to 2, and the total number of training epochs is set to 2,000 since they are denser. The dimension of node embeddings is set to 128 except on \dataset{Friendster}, where we use 96. For other hyperparameters, we follow their default values in previous works. For fair comparison, we report the training time of all methods with the same number of training epochs. We parallel the network augmentation in LINE. For DeepWalk, we store the random walks in memory, which is the fastest setting.

\subsection{Results on \dataset{Youtube}}

We first evaluate our hybrid system on the widely-used \dataset{Youtube} dataset. We compare the speed and performance of \Graphy with existing systems of node embedding. For existing systems, we replicate their parallel implementations and report their training time under the same number of training epochs. Table \ref{tab:time_youtube} presents the speed of different systems. Among all existing systems, LINE \cite{tang2015line} takes the minimal total time to run. However, the GPU implementation of LINE in OpenNE is even worse than its CPU counterpart, possibly due to the mini-batch SGD paradigm it uses. Compared to the current fastest system, LINE, \Graphy is much more efficient. With 4 GPUs, our system finishes training node emebdding on a million-scale network in only one and a half minutes. Even on a single GPU, \Graphy takes no more than 4 minutes and is still 19 times faster than LINE.

One may be curious about the performance of node embeddings learned by \Graphy. Therefore, we compare the performance of \Graphy with existing systems on the standard task of multi-label node classification. Note that normalizing the embeddings or not yields different trade off between Micro-F1 and Macro-F1 metrics. For fair comparison, we follow the practice in \cite{tang2015line} and train one-vs-rest linear classifiers over the normalized node embeddings. Table \ref{tab:performance_youtube} summarizes the performance over different percentages of training data. It is observed that \Graphy achieves the best or competitive results in most settings, showing that \Graphy does not sacrifice any performance. In some small percentage cases, \Graphy falls a little behind DeepWalk. This is because \Graphy uses negative sampling for optimization, while DeepWalk uses both hierarchical softmax and negative sampling, which could be more robust to few labeled data.

\subsection{Results on Larger Datasets}

\begin{table}[!h]
    \centering
    \begin{adjustbox}{max width=0.48\textwidth}
        \begin{tabular}{lccc}
            \toprule
            GraphVite   & \dataset{Friendster-small}    & \dataset{Hyperlink-PLD}   & \dataset{Friendster}  \\
            \midrule
            1 GPU       & 8.78 hrs                      & -                         & -                     \\
            4 GPU       & 2.79 hrs                      & 5.36 hrs                  & 20.3 hrs              \\
            \bottomrule
        \end{tabular}
    \end{adjustbox}
    \caption{Results of time on larger datasets. We only evaluate \dataset{Hyperlink-PLD} and \dataset{Friendster} with 4 GPUs since their embedding matrices cannot fit into the memory of a single GPU.}
    \label{tab:time_all}
\end{table}

\begin{figure*}
    \centering
    \begin{subfigure}{0.26\textwidth}
        \includegraphics[width=\textwidth]{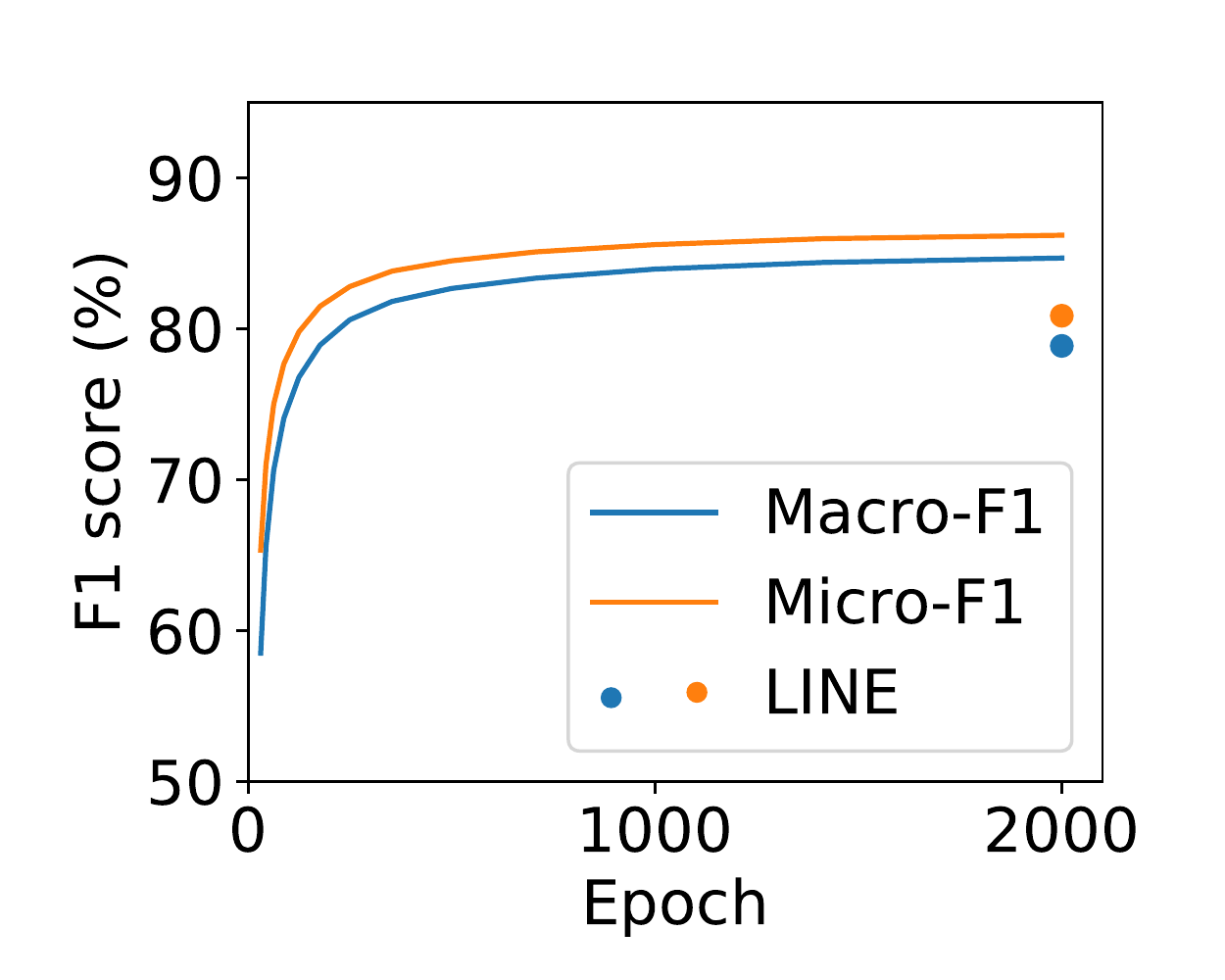}
        \caption{\dataset{Friendster-small}}
    \end{subfigure}
    \begin{subfigure}{0.26\textwidth}
        \includegraphics[width=\textwidth]{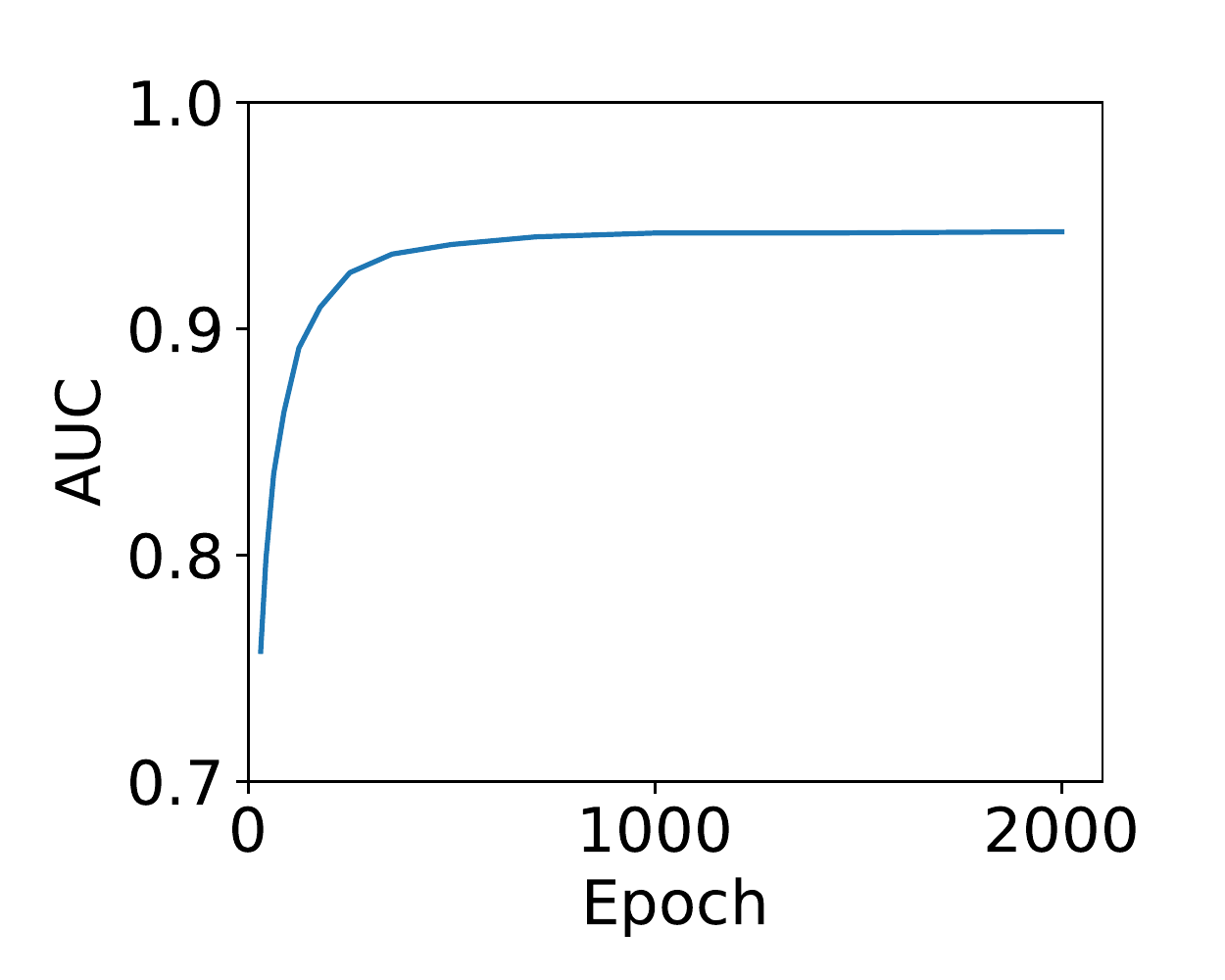}
        \caption{\dataset{Hyperlink-PLD}}
    \end{subfigure}
    \begin{subfigure}{0.26\textwidth}
        \includegraphics[width=\textwidth]{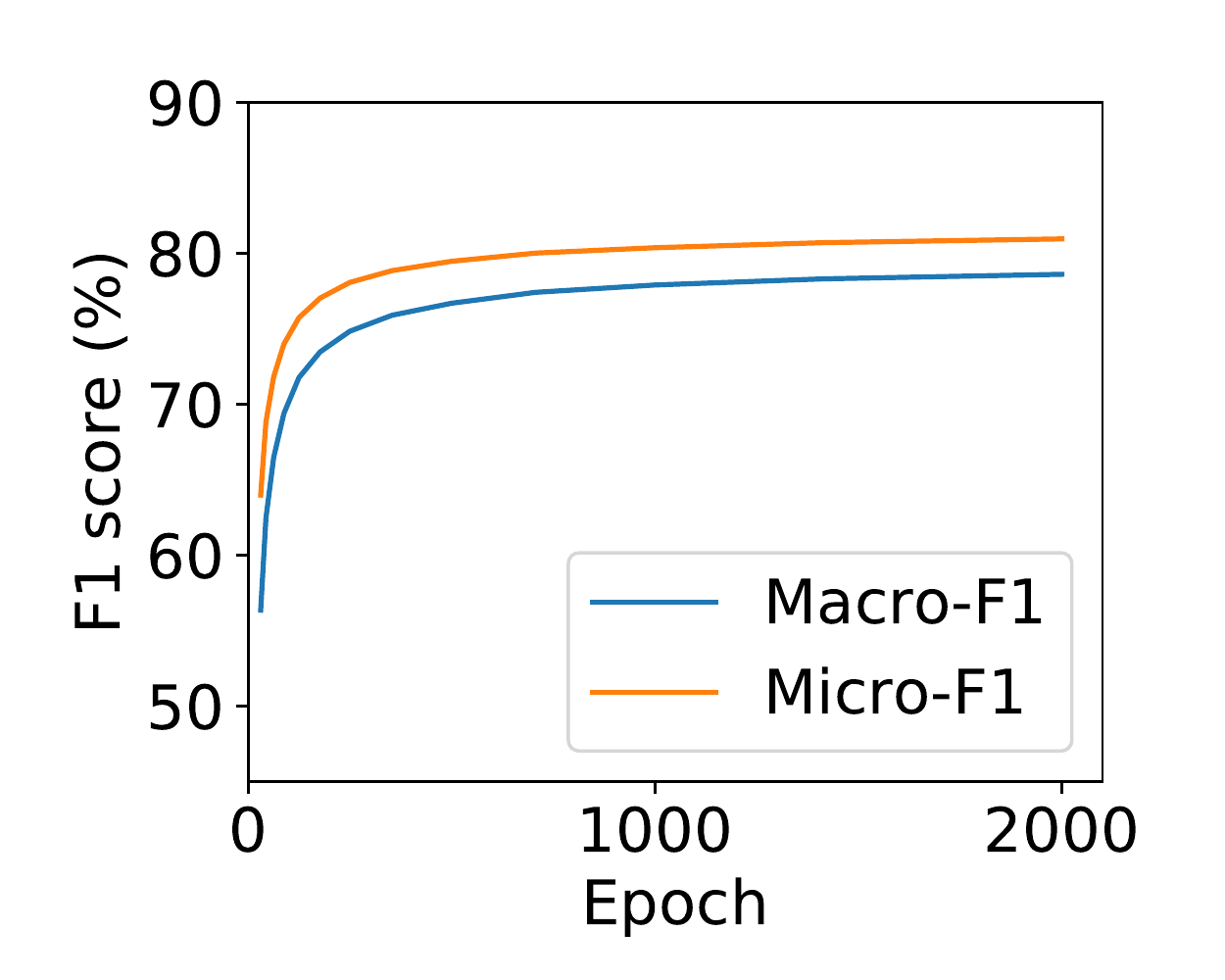}
        \caption{\dataset{Friendster}}
    \end{subfigure}
    \caption{Performance curves of \Graphy on larger datasets. For \dataset{Friendster}, we plot the results of LINE for reference. The other systems cannot solve any of these datasets within a week.}
    \label{fig:curves}
\end{figure*}

\begin{table*}
    \centering
    \begin{tabular}{lcccccc}
        \toprule
                            & Parallel Online       & Parallel Negative & Collaboration & \multirow{2}{*}{Micro-F1} & \multirow{2}{*}{Macro-F1} & \multirow{2}{*}{Training time}  \\
                            & Augmentation          & Sampling (4 GPUs) & Strategy  \\
        \midrule
        Single GPU baseline &                       &                   &               & 35.26     & 20.38     & 8.61 mins         \\
                            & \checkmark            &                   &               & 41.48     & 29.80     & 6.35 mins         \\
                            &                       & \checkmark        &               & 34.38     & 19.81     & 2.66 mins         \\
                            & \checkmark            & \checkmark        &               & 41.75     & 29.30     & 2.24 mins         \\
        \midrule
        GraphVite           & \checkmark            & \checkmark        & \checkmark    & 41.89     & 29.46     & \best{1.46 mins}  \\
        \bottomrule
    \end{tabular}
    \caption{Ablation of main components in \Graphy. Note that the baseline has the same GPU implementation with \Graphy and parallel edge sampling on CPU. The baseline should be regarded as a very strong one.}
    \label{tab:main_components}
\end{table*}

To demonstrate the scalability of \Graphy, we further test \Graphy on three larger networks. We learn the node embeddings of \dataset{Friendster-small} with 1 GPU and 4 GPUs. For \dataset{Hyperlink-PLD} and \dataset{Friendster}, since their embedding matrices cannot fit into the memory of a single GPU, we only evaluate them with 4 GPUs. Table \ref{tab:time_all} gives the training time of \Graphy on these datasets. The training time of baseline systems is not reported here, as all existing systems cannot solve such large networks in a week, except LINE \cite{tang2015line} on \dataset{Friendster-small}. Compared to them, \Graphy takes less than 1 day to train node embeddings on the largest dataset \dataset{Friendster} with 1.8 billion edges, showing that \Graphy can be an efficient tool for analyzing billion-scale networks.

We also evaluate the performance of the node embeddings on these datasets. For \dataset{Friendster-small} and \dataset{Friendster}, we test their node embeddings on multi-label node classification. The test set is built on the top-100 communities of \dataset{Friendster} and has a total of 39,679 nodes. We do not normalize the learned embeddings during evaluation, and report Macro-F1 and Micro-F1 based on 2\% labeled data. For \dataset{Hyperlink-PLD}, we adopt link prediction as the evaluation task since node labels are not available. We randomly exclude 0.01\% edges from the training set, and combine them with the same number of uniformly sampled negative edges to form a test set. Each edge sample is scored by the cosine similarity of two node embeddings. We report the AUC metric for link prediction. Figure \ref{fig:curves} presents the performance of \Graphy over different training epochs on these datasets. On \dataset{Friendster-small}, we also plot the performance of LINE for reference. Due to the long training time, we only report the performance of LINE by the end of all training epochs. It is observed that \Graphy converges on all these datasets. On the \dataset{Friendster-small} dataset, \Graphy significantly outperforms LINE. On the \dataset{Hyperlink-PLD}, we get an AUC of 0.943. On \dataset{Friendster}, the Micro-F1 reaches about 81.0\%. All the above observations verify the performance of our system.

%% file: 6_ablation.tex
\section{Ablation Study}
\label{sec:ablation}

To have a more comprehensive understanding of different components in \Graphy, we conduct several ablation experiments. For intuitive comparison, we evaluate these experiments on the standard \dataset{Youtube} dataset. We only report performance results based on 2\% labeled data due to space limitation. All the speedup ratios are computed with respect to LINE \cite{tang2015line}.

\subsection{What is the contribution of each main component?}

In the \Graphy, parallel online augmentation, parallel negative sampling, and the collaboration strategy are the main components in the sytem. Here we study how these components contribute to the performance of our system. We compare \Graphy with a strong baseline system with single GPU. Specifically, the baseline has the same GPU implementation as \Graphy, while it uses the standard parallel edge sampling instead of parallel online augmentation, and executes two stages sequentially.

Table \ref{tab:main_components} shows the results of this ablation. Compared to the baseline, we notice that parallel online augmentation helps improve the quality of node embeddings, since it introduces more connectivity to the sparse network. Besides, parallel online augmentation also accelerates the system a little, as it reuses nodes and reduces the amortized cost of each sample. With parallel negative sampling, we are able to employ multiple GPUs for training, and the speed is boosted by about 3 times. Moreover, the collaboration strategy even improves the speed and does not impact the performance.

\subsection{Is it necessary to perform pseudo shuffle?}

In parallel online augmentation, \Graphy performs pseudo shuffle to decorrelate the augmented edge samples, while some existing systems \cite{perozzi2014deepwalk, grover2016node2vec} do not shuffle their samples. We compare the proposed pseudo shuffle with three baselines, including no shuffle, a full random shuffle and an index mapping algorithm. The index mapping algorithm preprocesses a random mapping on the indexes of samples and saves the time of computing random variables.

Table \ref{tab:shuffle} gives the results of different shuffle algorithms on a single GPU. It is observed that all shuffle algorithms are about 1 percent better than the no shuffle baseline. However, different shuffle algorithms vary largely in their speed. Compared to the no shuffle baseline, the random shuffle and index mapping algorithms slow down the system by several times, while our pseudo shuffle has only a little overhead. Therefore, we conclude that pseudo shuffle is the best practice considering both speed and performance.

\begin{table}[!h]
    \centering
    \begin{tabular}{lcc}
        \toprule
        Shuffle algorithm   & Micro-F1(\%)  & Training time     \\
        \midrule
        None                & 40.41         & 3.60 mins         \\
        Random shuffle      & 41.61         & 17.1 mins         \\
        Index mapping       & 41.21         & 12.1 mins         \\
        \midrule
        Pseudo shuffle      & \best{41.52}  & \best{3.98 mins}  \\
        \bottomrule
    \end{tabular}
    \caption{Results of performance and speed by different shuffle algorithms. The proposed pseudo shuffle algorithm achieves the best trade off between performance and speed.}
    \label{tab:shuffle}
\end{table}

\subsection{What is a practical choice for episode size?}
\label{sec:exchangeability}

In parallel negative sampling, \Graphy relies on the property of gradient exchangeability to ensure its approximation to standard SGD. While the smaller episode size provides better exchangebility, it will increase the frequency of synchronization over rows of the embedding matrices, and thus slows down embedding training. To quantify such influence in speed and performance, we examine our system on 4 GPUs with different episode sizes. 

\begin{figure}[!h]
    \centering
    \begin{subfigure}{0.23\textwidth}
        \includegraphics[width=\textwidth]{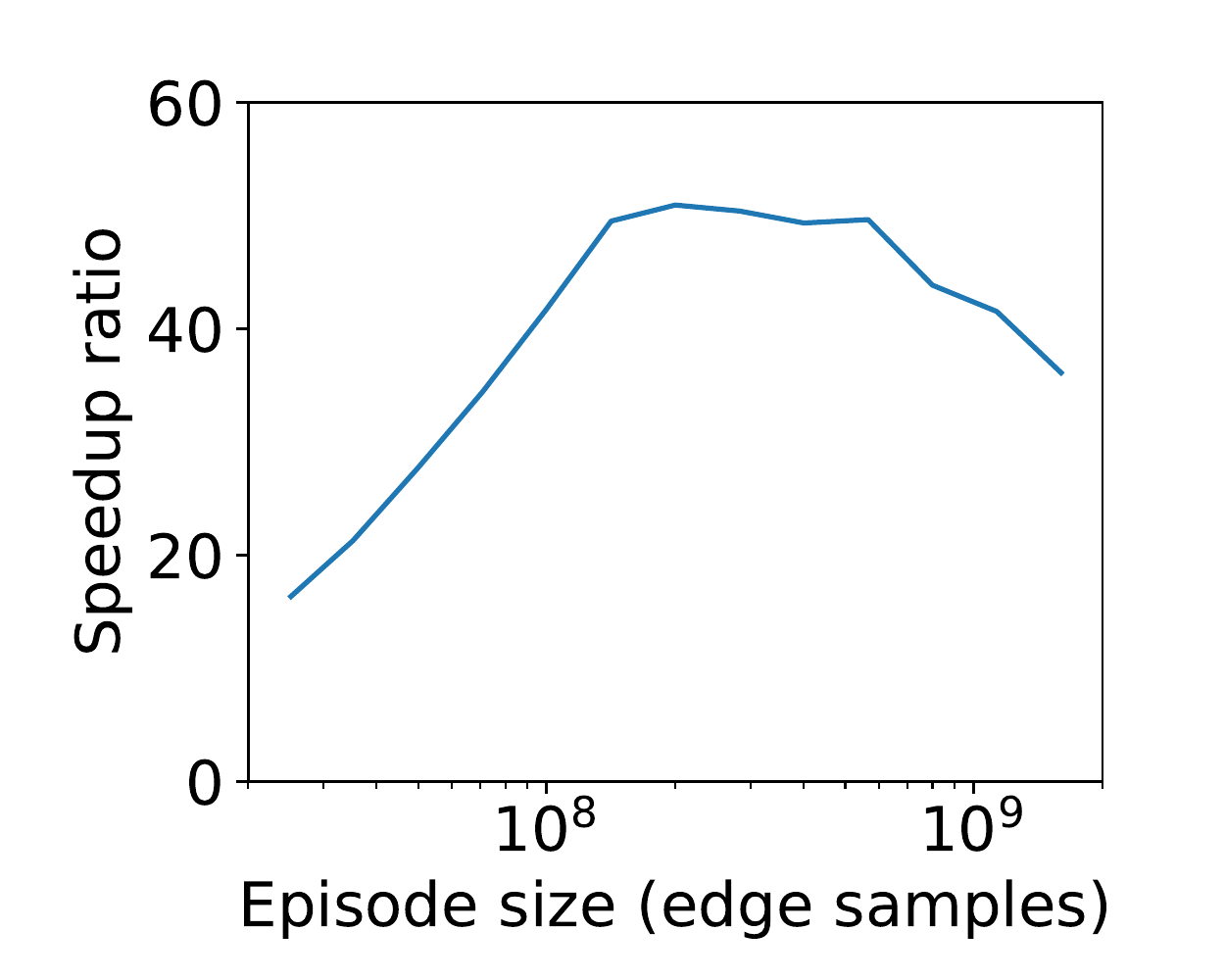}
    \end{subfigure}
    \begin{subfigure}{0.23\textwidth}
        \includegraphics[width=\textwidth]{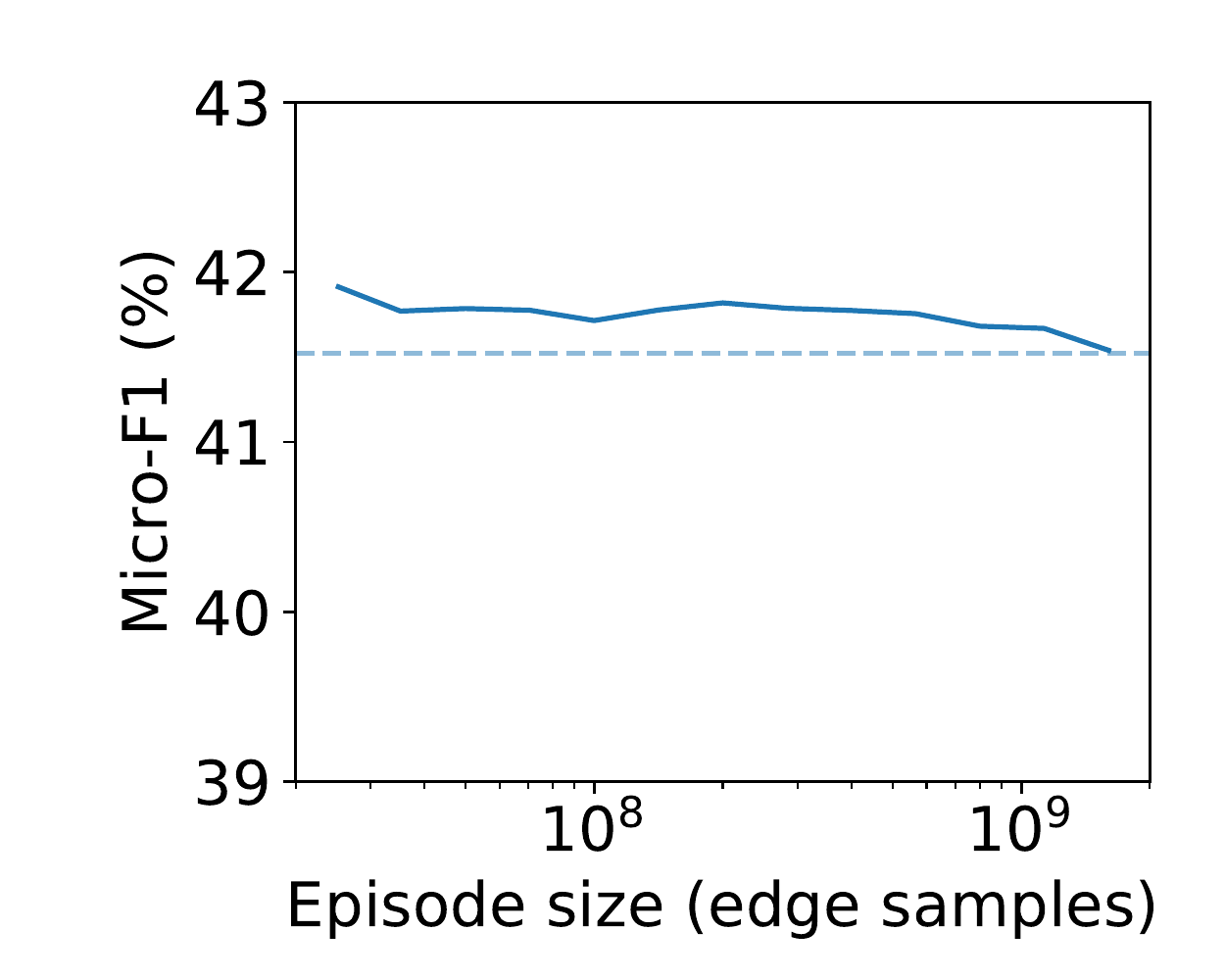}
    \end{subfigure}
    \caption{Speed and performance of \Graphy with respect to different episode sizes. The dashed line represents the single GPU baseline without parallel negative sampling.}
    \label{fig:episode_size}
\end{figure}

Figure \ref{fig:episode_size} plots the curves of speed and performance with respect to different episode sizes. On the performance side, we notice that the performance of \Graphy is insensitive to the choice of the episode size. Compared to the single GPU baseline, parallel negative sampling achieves competitive or slightly better results, probably due to the regularization effect introduced by partition. On the speed side, larger episode size achieves more speedup since it reduces the amortized burden of the bus. The speed drops at very large episode size, as there becomes only a few episodes in training. Therefore, we choose an episode size of $2*10^8$ edge samples for \dataset{Youtube}. Generally, the best episode size is proportional to $|V|$, so one can set the episode size for other networks accordingly.

\subsection{What is the speedup w.r.t the numebr of CPUs and GPUs?}

In \Graphy, both online augmentation and negative sampling can be parallelized on multiple CPUs or GPUs, and synchronization is only required between episodes. Therefore, our system should have great scalability. To verify that point, we investigate our system with different 
number of CPU and GPU. We change the number of GPU from 1 to 4, and vary the number of sampler per GPU from 1 to 5. The effective number of CPU threads is $\#GPU * (\#sampler~per~GPU + 1)$ as there is one scheduler thread for each GPU.

\begin{figure}[!h]
    \centering
    \includegraphics[width=0.3\textwidth]{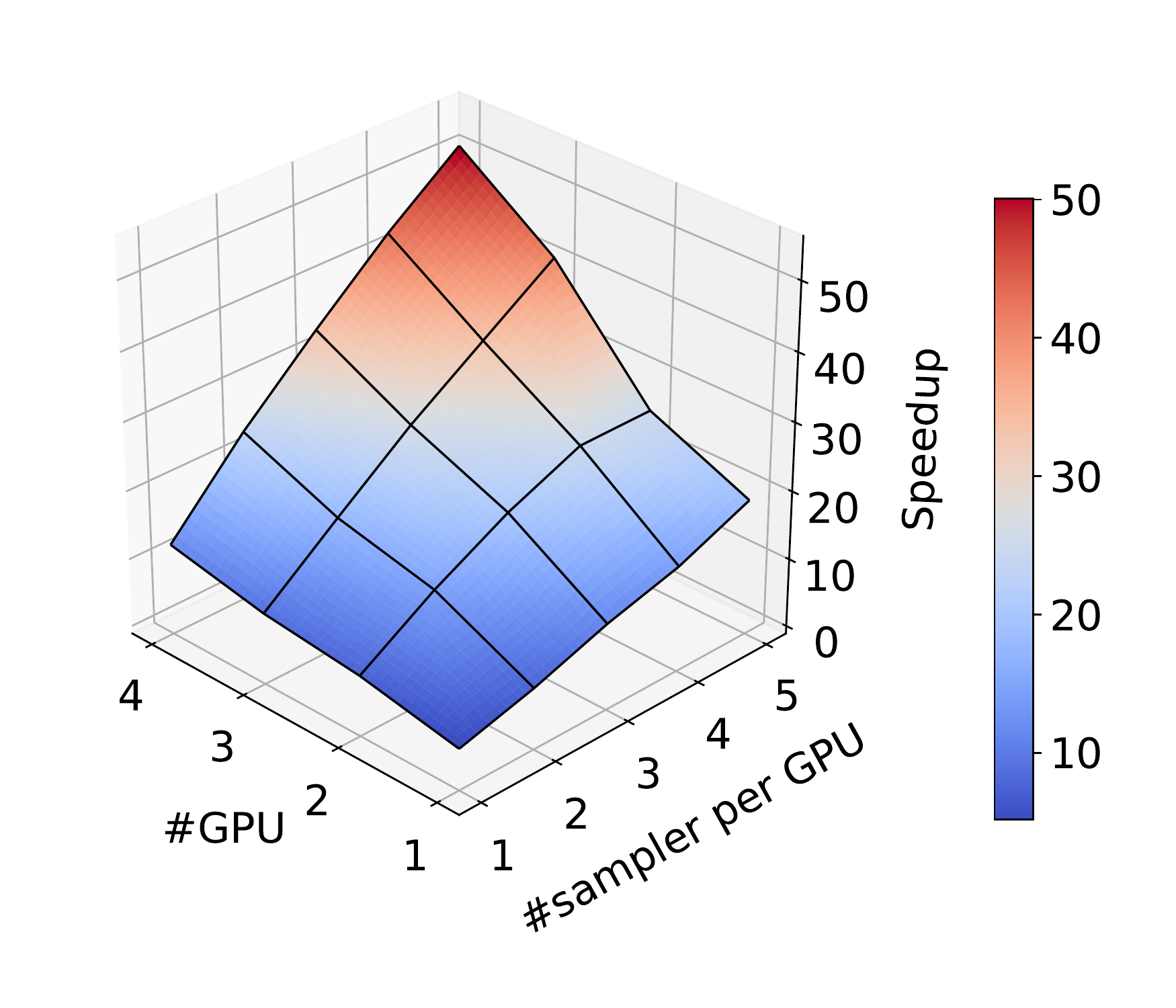}
    \caption{Results of speedup under different number of hardware. It is observed that the speedup is almost linear to the number of CPUs and GPUs.}
    \label{fig:scalability}
\end{figure}

Figure \ref{fig:scalability} plots the speedup ratio with respect to different number of CPUs and GPUs. The speedup ratio almost forms a plane over both variables, showing that our system scales almost linearly to the hardware. Quantitatively, \Graphy achieves a relative speedup of 11$\times$ when the hardware is scaled to 20$\times$. The speedup is about half of its theoretical maximum. We believe this is mainly due to the increased synchronization cost, as well as increased load on shared main memory and bus when we use more CPUs and GPUs.

\subsection{Does the hardware configuration matter?}

Up to now, all experiments are conducted on a server with Xeon E5 CPUs and Tesla P100 GPUs. One might wonder whether such a high performance depends on the specific hardware configuration. Therefore, we further test our system on an economic server with Core i7 CPUs and GTX 1080 GPUs.

Table \ref{tab:hardware} compares the results from two configurations. Different hardware does have difference in speed, but the gap is marginal. The time only increases to 1.6$\times$ when we move to the economic server. Note that this two configurations are almost the best and the worst in current machine learning servers, so one could expect a running time between these two configurations on his own hardware.

\begin{table}[!h]
    \centering
    \begin{tabular}{lcccc}
        \toprule
        Hardware                            & CPU threads   & GPU   & Training time \\
        \midrule
        \multirow{2}{*}{Tesla P100 server}  & 6             & 1     & 3.98 mins     \\
                                            & 24            & 4     & 1.46 mins     \\
        \midrule
        \multirow{2}{*}{GTX 1080 server}    & 3             & 1     & 6.28 mins     \\
                                            & 12            & 4     & 2.48 mins     \\
        \bottomrule
    \end{tabular}
    \caption{Training time of \Graphy under different hardware configurations. Generally \Graphy may take a time between these two configurations on most hardware.}
    \label{tab:hardware}
\end{table}

%% file: 2_related_work.tex
\section{Related Work}
\label{sec:related}

Node embedding has been proven effective in a wide range of applications, such as node classification \cite{hamilton2017representation}, link prediction \cite{liben2007link}, and network visualization \cite{tang2016visualizing}. Many different methods \cite{perozzi2014deepwalk, tang2015line, cao2015grarep, grover2016node2vec, wang2016structural, qiu2018network, tsitsulin2018verse, velivckovic2018deep} have been proposed to learn node embeddings that preserve the structure of networks from different aspects. Among them, DeepWalk \cite{perozzi2014deepwalk}, LINE \cite{tang2015line} node2vec \cite{grover2016node2vec} and VERSE \cite{tsitsulin2018verse} are built on either edge or path samples of networks, which makes them the most scalable methods of all. Our work follows this stream and is related to these methods.

\medskip \noindent \textbf{Node Embedding Algorithm}
Generally, node embedding algorithms consist of two stages, namely network augmentation and embedding training. The network augmentation stage is widely adopted in existing methods \cite{perozzi2014deepwalk, tang2015line, cao2015grarep, grover2016node2vec, qiu2018network} to improve the performance of learned embeddings on sparse networks. DeepWalk \cite{perozzi2014deepwalk} and node2vec \cite{grover2016node2vec} augment networks by generating random paths according to different distributions. The edge samples derived by path are correlated in those methods. LINE \cite{tang2015line} directly adds edges to networks and generates independent edge samples. GraRep \cite{cao2015grarep} and NetMF \cite{qiu2018network} take different powers of the adjacency matrix as augmentation. Our parallel online augmentation generates decorrelated edge samples using pseudo shuffle, and thus is close to the augmentation in LINE. However, our augmentation does not need to store the whole augmented network, which saves a lot of disk and memory usage compared to existing methods.

In the embedding training stage, most existing node embedding algorithms \cite{perozzi2014deepwalk, tang2015line, grover2016node2vec} train node embedding with standard negative sampling \cite{mikolov2013distributed} in a shared memory space. While there is a parallel word embedding algorithm \cite{stergiou2017distributed} that restricts negative sampling within the context partition of each worker, it still needs to transfer rows of embedding matrices between each worker for positive samples. By contrast, our parallel negative sampling trains on orthogonal sample blocks and does not need any transfer between worker during an episode. The most related method is the distributed SGD used in large-scale matrix factorization algorithms \cite{gemulla2011large, zhuang2013fast, yun2014nomad, bhavana2019bmf}. These methods divide the input matrix into $n \times n$ blocks and factorize orthogonal blocks simultaneously. Different from these methods, our system mainly focuses on the negative sampling technique and is designed for the node embedding task.

\medskip \noindent \textbf{Node Embedding System}
From the perspective of system, our work belongs to the parallel implementation of node embedding. There are many CPU parallel systems, including DeepWalk \textsuperscript{\ref{fn:deepwalk_url}}, LINE \textsuperscript{\ref{fn:line_url}}, node2vec \textsuperscript{\ref{fn:node2vec_url}} and VERSE \footnote{\url{https://github.com/xgfs/verse}}. These systems use asynchronous SGD \cite{recht2011hogwild} in embedding training and exploit multiple CPU threads for acceleration. Due to the limited computation speed of CPUs, such systems cannot scale to ten-million-scale networks without a large CPU cluster. Recently, there are some GPU parallel systems \cite{thunlp2017openne} built on deep learning frameworks like TensorFlow \cite{abadi2016tensorflow} or PyTorch \cite{paszke2017automatic}. Since existing frameworks are based on mini-batch SGD paradigm, these systems severely suffer the problem of limited bus bandwidth, and are even worse than their CPU counterparts. Compared to them, \Graphy is a hybrid CPU-GPU system that leverages distinct advantages of CPUs and GPUs, and uses them collaboratively to train node embedding, which makes it much faster than either pure CPU or mini-batch-SGD based systems.

In addition, parallel word embedding systems \cite{mikolov2013distributed, ji2016parallelizing, simonton2017efficient, gupta2017blazingtext} are also very related to our work, since they share similar embedding training and negative sampling steps with node embedding. Among these methods, Wombat \cite{simonton2017efficient} and BlazingText \cite{gupta2017blazingtext} accelerate training with GPUs. Wombat only supports single GPU. BlazingText can scale to multiple GPUs, but it simply makes a copy of the parameter matrices on each GPU. Our system is more efficient than BlazingText in two aspects. First, our system partition the parameter matrices and consumes less memory on each GPU. Second, our system requires less synchronization cost, as GPUs do not share any rows in the parameter matrices.

%% file: 7_conclusion.tex
\section{Conclusion}
\label{sec:conclusion}

In this paper, we present a high-performance CPU-GPU hybrid system for node embedding. Our system extends existing node embedding methods to GPUs and significantly accelerates training node embeddings on a single machine. With parallel online augmentation, \Graphy efficiently utilizes CPU threads to generate augmented edge samples for node embedding training. With parallel negative sampling, \Graphy enables training node embeddings on multiple GPUs without much synchronization. A collaboration strategy is also developed to reduce the synchronization cost between CPUs and GPUs. Experiments on 4 large networks prove that \Graphy significantly outperforms existing systems in speed without sacrifice on performance. In the future, we plan to generalize our system to semi-supervised settings and graph neural networks, such as graph convolutional networks \cite{kipf2016semi}, graph attention networks \cite{velickovic2017graph}, and neural message passing networks \cite{gilmer2017neural}.

\nocite{newman2008distributed}

%% file: 8_acknowledgement.tex
\section*{Acknowledgements}

We would like to thank Compute Canada \footnote{\url{https://www.computecanada.ca}} for supporting GPU servers. Jian Tang is supported by the Natural Sciences and Engineering Research Council of Canada and the Canada CIFAR AI Chair Program. We specially thank Wenbin Hou for useful discussions on C++ and GPU programming techniques, and Sahith Dambekodi for proofreading this paper.